\documentclass{article}

\usepackage{arxiv}

\usepackage[utf8]{inputenc}
\usepackage[fleqn]{amsmath}
\usepackage{amsfonts}
\usepackage{graphicx}
\usepackage{tikz}
\usetikzlibrary{shapes.geometric, arrows.meta, positioning}
\usepackage{natbib}
\usepackage{geometry}
\usepackage[hidelinks]{hyperref}
\usepackage{lineno}
\usepackage{caption}
\usepackage{subcaption}
\usepackage{setspace}
\usepackage{xcolor}
\usepackage{float}
\usepackage{algorithm}
\usepackage{algpseudocode}
\usepackage{comment}
\usetikzlibrary{calc}

\title{Real Time Headway Predictions in Urban Rail Systems and Implications for Service Control: A Deep Learning Approach}

\author{
  Muhammad Usama \\
  Department of Civil and Environmental Engineering\\
 Northeastern University\\
  Boston, MA, 02115 \\
  \texttt{m.usama@northeastern.edu, engrmusama90@gmail.com} \\
   \And
 Haris N. Koutsopoulos \\
  Department of Civil and Environmental Engineering\\
 Northeastern University\\
  Boston, MA, 02115 \\
  \texttt{h.koutsopoulos@northeastern.edu}\\
}

\begin{document}
\maketitle
\begin{abstract}
Efficient real-time dispatching in urban metro systems is essential for ensuring service reliability, maximizing resource utilization, and improving passenger satisfaction. This study presents a novel deep learning framework centered on a Convolutional Long Short-Term Memory (ConvLSTM) model designed to predict the complex spatiotemporal propagation of train headways across an entire metro line. By directly incorporating planned terminal headways as a critical input alongside historical headway data, the proposed model accurately forecasts future headway dynamics, effectively capturing both their temporal evolution and spatial dependencies across all stations. This capability empowers dispatchers to evaluate the impact of various terminal headway control decisions without resorting to computationally intensive simulations. We introduce a flexible methodology to simulate diverse dispatcher strategies, ranging from maintaining even headways to implementing custom patterns derived from observed terminal departures. In contrast to existing research primarily focused on passenger load predictioning or atypical disruption scenarios, our approach emphasizes proactive operational control. Evaluated on a large-scale dataset from an urban metro line, the proposed ConvLSTM model demonstrates promising headway predictions, offering actionable insights for real-time decision-making. This framework provides rail operators with a powerful, computationally efficient tool to optimize dispatching strategies, thereby significantly improving service consistency and passenger satisfaction.
\end{abstract}


\section{Introduction}

The rapid growth of urban populations demands sustainable and efficient transportation solutions to alleviate congestion and reduce environmental impacts. In response, mass transit systems have become indispensable for facilitating high-capacity, low-emission mobility in densely populated areas. These metro systems serve as the backbone of urban transportation infrastructure, offering efficient service that supports the daily movement of millions.

However, despite their vital role, metro systems increasingly struggle to maintain service quality. Challenges such as overcrowding, aging infrastructure, and operational inefficiencies are growing more prevalent \cite{pojani2015sustainable, vuchic2005urban, tumlin2011sustainable}. The quality of service, particularly during peak hours, is significantly impacted by factors such as passenger load, train punctuality, and, crucially, operational consistency, which is often reflected in train headways \cite{chaowen2019, AN2020102100}.
Consequently, accurate and real-time prediction of service quality indicators is critical for ensuring reliable urban rail operations. Such predictions enable operators to anticipate demand fluctuations, enhance service levels, allocate resources efficiently, and respond proactively to potential disruptions. At the same time, commuters benefit from timely and robust information on schedules and comfort, allowing them to plan journeys with greater confidence \cite{kermHaris2024}.

Traditional approaches for short-term prediction problems in transportation, which often involve nonlinear inference with inherent spatial and temporal dependencies, have relied on a range of methodologies. These include classical statistical methods such as the autoregressive integrated moving average method \cite{Guang2018, dong2011joint, emami2020short,xu2017real}, various machine learning models like support vector machines \cite{xiao2012traffic, sun2014use, wu2004travel} and random forests \cite{Feng2023}.

Effective real-time dispatching in urban metro systems is paramount for maintaining service reliability, optimizing train and operator utilization, and ultimately enhancing passenger experience \cite{wang2022real, shen2001optimal, flamini2008real, d2009advanced, fabian2018improving, berrebi2018translating}. Current dispatching decisions, especially at terminal stations, have cascading effects on subsequent headways across the entire line, yet dispatchers often lack immediate, quantifiable insights into these downstream impacts \cite{wolofsky2019towards, zhou2020evaluation, yousefi2025headway, soza2019lessons}. These challenges are compounded by the complexity of urban metro networks, where delays at one point can propagate system-wide, leading to overcrowding and reduced service quality \cite{vuchic2005urban, saidi2023train}. Therefore, a comprehensive predictive approach for a public transport metro line must effectively address several critical aspects: (i) interactions among all trains moving along the line, (ii) robustness under critical situations such as minor disruptions and varying operational conditions, and most importantly, (iii) providing dispatchers with real-time, actionable insights into the downstream effects of their control decisions, particularly regarding terminal headways, without requiring extensive, time-consuming simulations.

The key contributions of this paper are summarized as follows:
\begin{itemize}
    \item We propose a novel ConvLSTM-based deep learning framework tailored for the spatiotemporal prediction of train headways across an entire metro line, effectively modeling both temporal evolution and spatial propagation of headway dynamics..
    \item We introduce an approach to incorporate target terminal headways as direct model inputs, enabling prediction of downstream headway impacts from dispatcher decisions, addressing an important gap in existing models.
    \item We develop a computationally efficient methodology for simulating diverse dispatcher strategies, including target headway adherence and custom departure patterns, facilitating real-time operational planning and evaluation.
\end{itemize}


\section{Methodology}

This section presents the methodology for developing a convLSTM-based deep learning framework designed to predict train headways across an urban metro line. A key innovation of this approach is the explicit incorporation of target terminal headways as an input, which allows the model to capture and predict the downstream impacts of dispatcher decisions on real-time operational dynamics. The framework leverages spatiotemporal modeling to address the critical challenges of service reliability and operational efficiency in metro systems.

\subsection{Data Requirements and Preprocessing}
\label{sec:data_preprocessing}

The proposed methodology uses Automatic Vehicle Location (AVL) system. This system generates detailed records of individual train events, specifically the times at which each signal block is activated by a train. These activation times form train trajectories, which capture train movements across track circuits. To transform raw event-based data into a structured format suitable for spatiotemporal deep learning, a multi-stage preprocessing pipeline is employed, as detailed in Algorithm \ref{alg:data_preprocessing}.

\begin{algorithm}[htbp]
  \caption{Data Preprocessing}
  \label{alg:data_preprocessing}
  \begin{algorithmic}[1]
    \Require Raw AVL data.
    \Ensure Historical input tensor $X \in \mathbb{R}^{L \times N_d \times N_{\text{dir}} \times 1}$, terminal input tensor $T \in \mathbb{R}^{F \times N_{\text{dir}} \times 1}$, and target output tensor $Y \in \mathbb{R}^{F \times N_d \times N_{\text{dir}} \times 1}$.
    \State Initialize preprocessing parameters: time bin size $\Delta T$, number of distance bins $N_d$, minimum distance $d_{\min}$, maximum distance $d_{\max}$, look-back window $L$, and prediction horizon $F$.
    \For{each day AVL dataset}
      \State Compute the time bin index $i$ using Equation \ref{eq:time_bin}.
      \State Compute the distance bin index $j$ using Equation \ref{eq:distance_bin}.
      \State Assign the event to the corresponding grid cell $(i, j, k)$.
    \EndFor
    \For{each grid cell $(t, j, k)$}
      \State Compute $H(t, j, k)$ as the mean of headway for events in the cell, using Equation \ref{eq:headway_calc}.
    \EndFor
    \State Normalize $H(t, j, k)$ to the interval $[0, 1]$.
    \For{each unique combination of day and direction}
      \State Get input tensor $X$ over $L$ preceding time steps, as defined in Equation \ref{eq:input_tensor}.
      \State Get the future terminal input tensor $T$ over $F$ prediction steps, as defined in Equation \ref{eq:terminal_input_tensor}.
      \State Get the target output tensor $Y$ over $F$ prediction steps, as defined in Equation \ref{eq:target_output_tensor}.
    \EndFor
    \State \Return $X$, $T$, and $Y$
  \end{algorithmic}
\end{algorithm}

The preprocessing pipeline discretizes the continuous time and distance dimensions of metro line operations into fixed-size bins, creating a grid-like representation of headway dynamics. For an operational time window $[t_{\text{start}}, t_{\text{end}}]$, time is segmented into bins of uniform size $\Delta T$. The time bin index for an event occurring at time $t$ is defined as:

\begin{linenomath}
  \begin{equation}
   \quad \quad \text{time\_bin}_i = \left\lfloor \frac{t - t_{\text{start}}}{\Delta T} \right\rfloor, \quad t \in [t_{\text{start}}, t_{\text{end}})
    \label{eq:time_bin}
  \end{equation}
\end{linenomath}

Similarly, the track length of the metro line is divided into $N_d$ equally sized distance bins, each representing a fixed-length segment. For an event at distance $d$ from the line’s origin, the distance bin index is given by:

\begin{linenomath}
  \begin{equation}
    \quad \quad\text{distance\_bin}_j = \left\lfloor \frac{d - d_{\min}}{(d_{\max} - d_{\min}) / N_d} \right\rfloor, \quad j \in \{0, 1, \ldots, N_d-1\}
    \label{eq:distance_bin}
  \end{equation}
\end{linenomath}

\begin{figure}[htbp]
  \centering
  \includegraphics[width=0.5\linewidth]{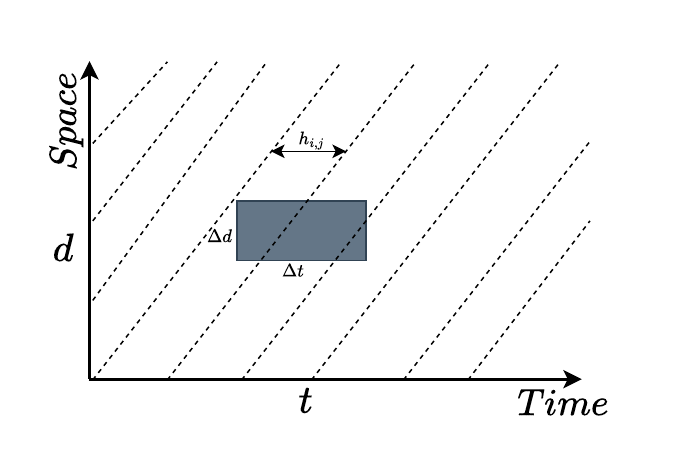}
  \caption{Train trajectories are plotted with time ($t$) on the x-axis and distance ($d$) on the y-axis, illustrating the spatio-temporal locations of signal block activations with headways, discretized into grid cells of size $\Delta T \times \Delta d$.}
  \label{fig:trajectory_headway}
\end{figure}

For each grid cell $(t, j, k)$, where $t$ is the time bin index, $j$ is the distance bin index, and $k$ is the direction index, the headway $H(t, j, k)$ is computed as the average of all individual headway values associated with train events within that cell, as shown in Figure \ref{fig:trajectory_headway}:

\begin{linenomath}
  \begin{flalign}
    \quad \quad &H(t, j, k) = \frac{1}{N_{t,j,k}} \sum_{i \in S_{t,j,k}} h_{i} \label{eq:headway_calc}
  \end{flalign}
\end{linenomath}
where $S_{t,j,k}$ denotes the set of trajectory points within the grid cell, $h_{i}$ is the headway of the $i$-th trajectory point, and $N_{t,j,k}$ is the number of points in the cell. For grid cells with no observed events ($N_{t,j,k} = 0$), missing headway values are imputed based on a logic of headway observed by a station.

The final stage of preprocessing involves structuring the spatio-temporal headway data into input-output sequences for deep learning. For each unique combination of replication and direction, time-series sequences are extracted. The historical input tensor $X$ captures headway patterns over $L$ preceding time steps:

\begin{linenomath}
  \begin{equation}
    \quad \quad X = \big[H(t-L, j, k), \ldots, H(t-1, j, k)\big] \in \mathbb{R}^{L \times N_d \times N_{\text{dir}} \times 1}
    \label{eq:input_tensor}
  \end{equation}
\end{linenomath}

where $L$ represents the historical look-back window, $N_d$ is the number of distance bins, and $N_{\text{dir}}$ accounts for directions (e.g., Northbound and Southbound). The future terminal headway input tensor $T$ incorporates Target headways at terminal stations over $F$ prediction steps:

\begin{linenomath}
  \begin{equation}
    \quad \quad T = \big[H(t, s, k), \ldots, H(t+F-1, s, k)\big] \in \mathbb{R}^{F \times N_{\text{dir}} \times 1}
    \label{eq:terminal_input_tensor}
  \end{equation}
\end{linenomath}

This tensor enables the model to leverage dispatcher decisions or predefined schedules at terminal stations, enhancing prediction accuracy. The output tensor $Y$ represents the actual future headways to be predicted over $F$ time steps across all distance bins:

\begin{linenomath}
  \begin{equation}
    \quad \quad Y = \big[H(t, j, k), \ldots, H(t+F-1, j, k)\big] \in \mathbb{R}^{F \times N_d \times N_{\text{dir}} \times 1}
    \label{eq:target_output_tensor}
  \end{equation}
\end{linenomath}

\subsection{ConvLSTM Model Architecture}
\label{subsec:model_architecture}
\begin{figure}
    \centering
    \includegraphics[width=\linewidth]{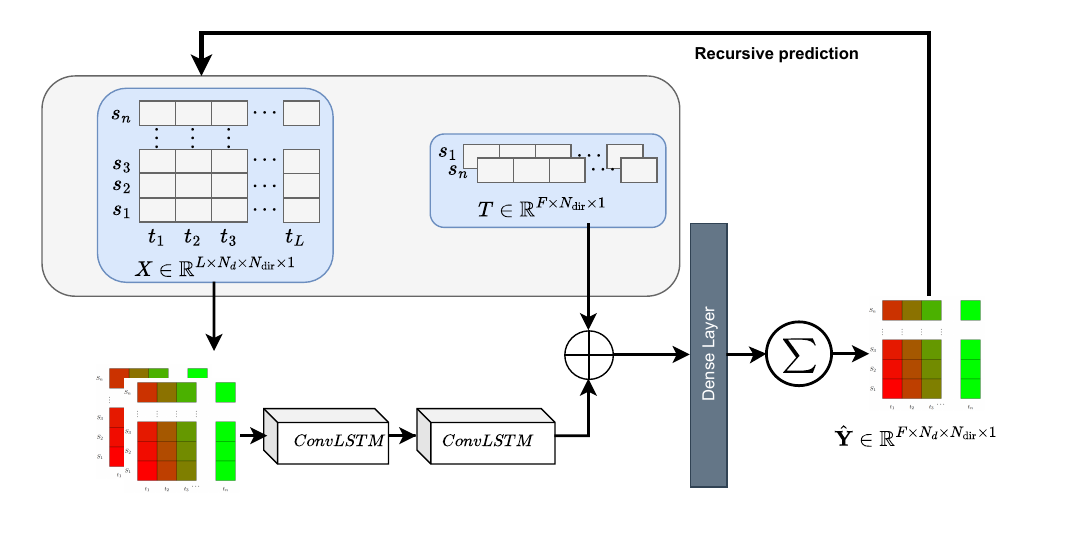}
    \caption{ConvLSTM-based model architecture for spatiotemporal headway prediction, integrating historical and planned terminal headway inputs.}
    \label{fig:model_architecture}
\end{figure}

The proposed deep learning framework is built upon a hybrid ConvLSTM architecture, designed to effectively capture both the spatial correlations of headways across the metro line and their complex temporal dependencies over time. Unlike conventional LSTM networks, which primarily process sequential data without explicit spatial consideration, ConvLSTM integrates convolutional operations directly within its recurrent cells. This unique integration allows the model to learn hierarchical spatial features directly from the spatiotemporal data. The complete model architecture, providing a visual representation of its components and data flow, is illustrated in Figure \ref{fig:model_architecture}.

\begin{equation}
\quad \quad \text{MSE} = \frac{1}{N}\sum_{i=1}^{N} (Y_i - \hat{Y}_i)^2
\label{eq:mse}
\end{equation}
where $Y_i$ is the actual headway value, $\hat{Y}_i$ is the predicted headway value, and $N$ is the total number of predictions.

Finally, the model's predictive performance is assessed through a performance evaluation process on a validation set. Key quantitative metrics, including MSE, Root Mean Squared Error (RMSE), and the coefficient of determination ($R^2$), are computed to provide an assessment of overall accuracy and goodness-of-fit.


\section{Application}

This section presents the evaluation of the proposed ConvLSTM framework for spatiotemporal headway prediction in urban metro systems. We first detail the experimental setup, including data preparation and model configuration. Subsequently, we present the overall predictive performance, followed by an in-depth analysis of headway prediction accuracy across various prediction horizons and at different stations along the metro line. The discussion then elaborates on the implications of integrating planned terminal headways and the model's capacity to support proactive dispatching strategies.

\subsection{Experimental Setup and Model Configuration}

\begin{figure}
    \centering
    \includegraphics[width=0.7\linewidth]{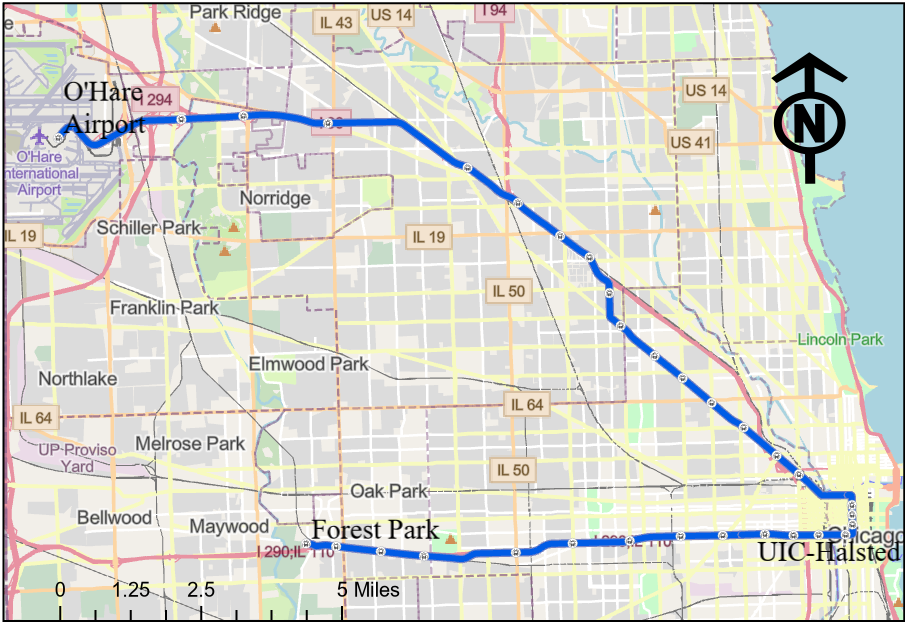}
    \caption{Map of CTA Blue Line with terminals at Forest Park and O'Hare and UIC-Halsted serve as short-turning station}
    \label{fig:cta_bl_map}
\end{figure}

The Chicago Transit Authority (CTA) Blue Line, a vital artery in Chicago's urban rail network, serves as testbed for our proposed spatio-temporal headway prediction framework. The map of the CTA Blue Line is shown in Figure \ref{fig:cta_bl_map}. This line presents an challenging operational environment, primarily due to inherent high passenger demands, but more significantly, by its complex short-turning operations. During peak hours, an operational strategy is deployed where a substantial proportion of southbound trains, instead of continuing to the Forest Park Terminal, execute a scheduled short turn at UIC-Halsted station, reversing course to become northbound trains. This maneuver is essential to maintain high frequency and meet the intense passenger demand in the high-demand section spanning between O'Hare International Airport and UIC-Halsted. Despite the careful scheduling, the short-turning trains often conflict with full-service northbound trains dispatched from the Forest Park Terminal due to uncertainty in operations and demand. The current dispatching protocols from the Forest Park Terminal often operate in isolation, without fully accounting for the real-time or future impact of short-turning trains at UIC-Halsted, leading to unpredictable interactions and headway variability. This dynamic interplay, where planned short turns conflict with full-service dispatches, creates a complex system, making accurate headway prediction an important tool for making dispatching decisions.

To address the complexities and data quality issues associated with currently available AVL data of CTA Blue Line, we opted to use a high-resolution dataset generated from a calibrated microsimulation model \cite{yousefi2025headway}, developed and validated for the CTA Blue Line, as a proof of concept. To ensure a sufficient volume of data and capture a wide range of operational dynamics, we ran 50 replications of the model with random initializations for the afternoon peak period, specifically from 15:30 to 18:00. The resulting simulated data represents the metro line's underlying infrastructure and train movements, providing a clean data for the spatiotemporal headway matrices required by the proposed framework.



For this study, the entire track was divided into 64 equally sized distance bins about 2200 feet long. A single distance bin may contain several smaller track sections depending on the length of segments. The raw train movement data is transformed into headway observations and aggregated into 1-minute time bins. This created a structured time × distance grid of headway values for both the Northbound and Southbound directions.

\begin{figure}[H]
\centering
\begin{subfigure}{0.48\textwidth}
    \centering
    \includegraphics[width=\textwidth]{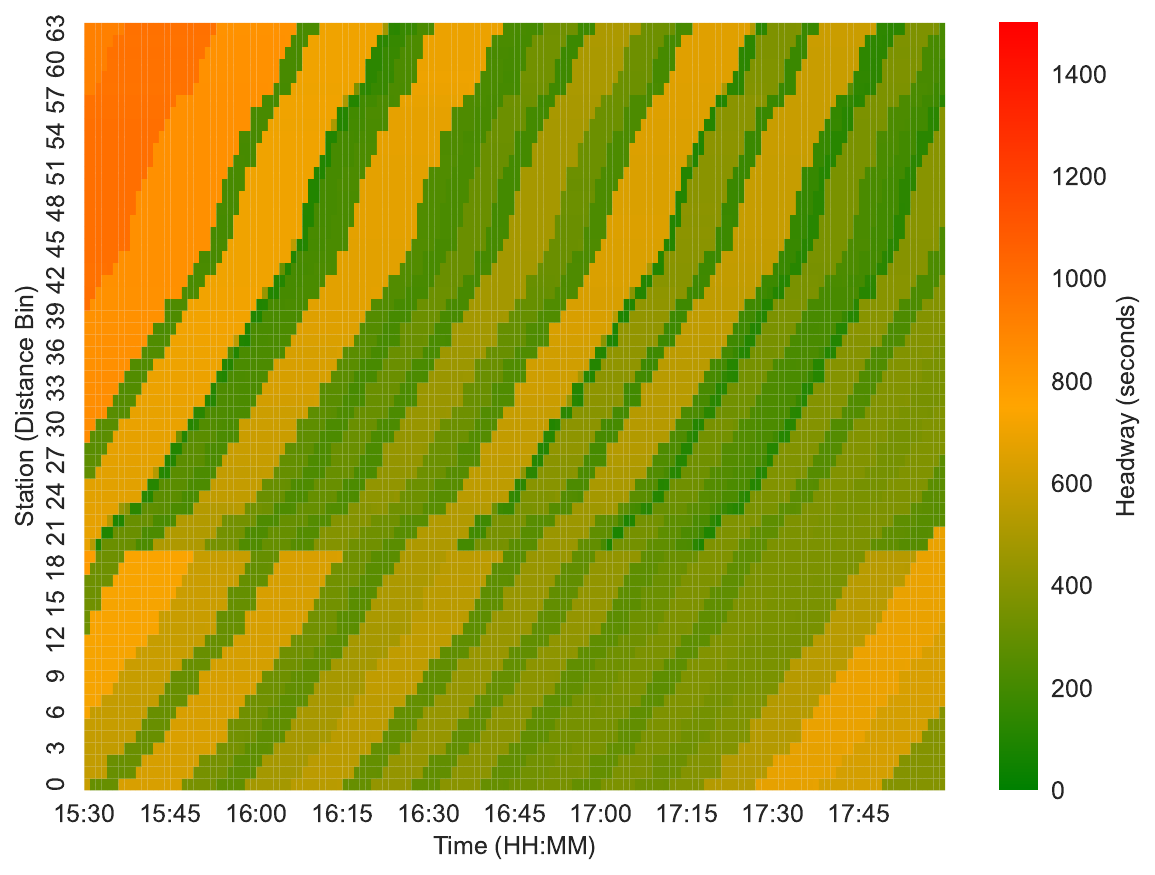}
    \caption{Northbound}
    \label{fig:northbound_heatmap}
\end{subfigure}
\hfill
\begin{subfigure}{0.48\textwidth}
    \centering
    \includegraphics[width=\textwidth]{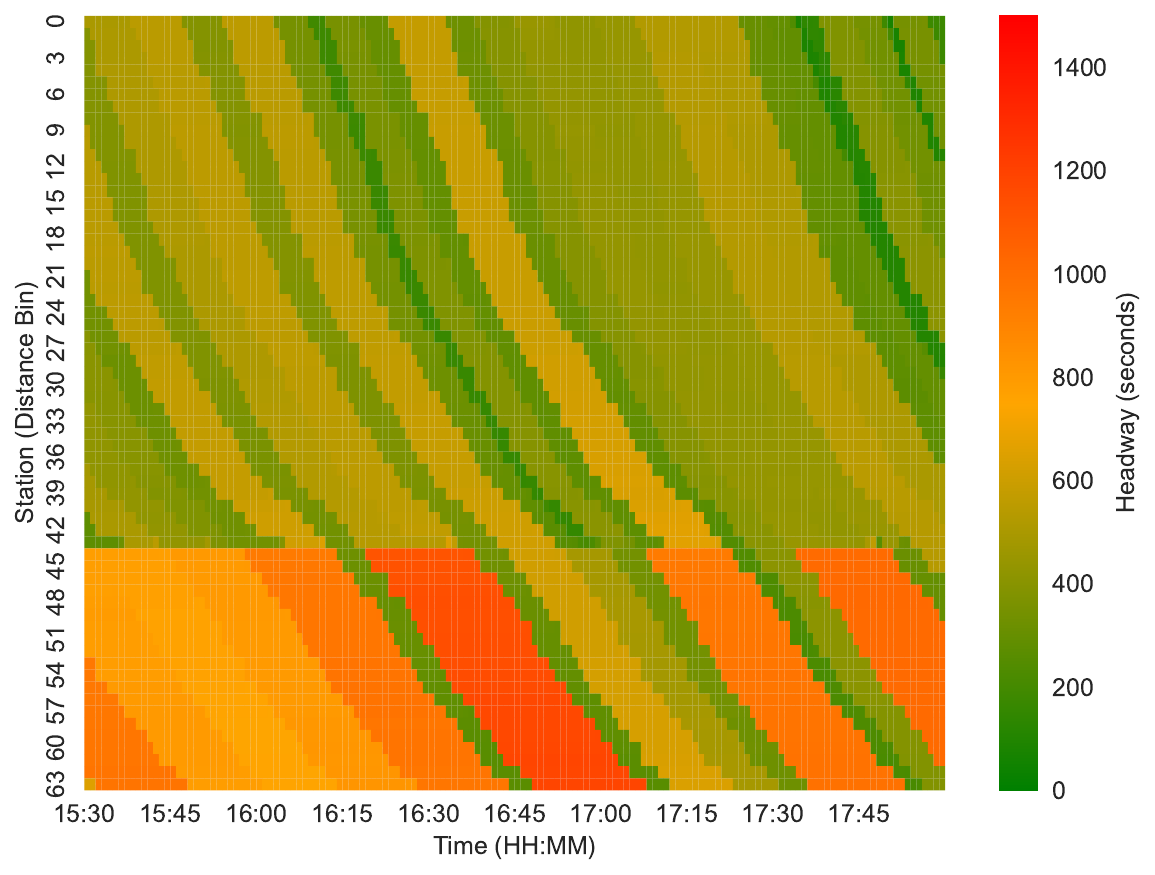}
    \caption{Southbound}
    \label{fig:southbound_heatmap}
\end{subfigure}
\caption{Headway heatmaps for the Blue Line during the afternoon peak period (15:30--18:00). The plots illustrate headway values (in seconds) across 64 distance bins and 1-minute time bins. The headway changes at distance bin 21 indicate the short-turning station UIC-Halsted, where southbound trains switch to northbound, leading to a complex operational situation.}
\label{fig:data_headway_heatmaps}
\end{figure}

This spatiotemporal micrograph, as shown in Figure \ref{fig:data_headway_heatmaps}, provides a visualization of headway dynamics across the CTA Blue Line. The heatmaps are structured with time on the x-axis, and distance bins on the y-axis. To reflect the operational flow, the y-axis for the Northbound direction begins at the Forest Park terminal, while the y-axis for the Southbound direction starts from the O'Hare terminal. The deliberate inversion of the y-axis for Southbound trains allow for better representation of headway propagation from top to bottom. The color gradient, transitioning from green (smaller headways) to red (longer headways), highlights headway variations across the line. The micrographs show operational complexities, notably at the short-turning station UIC-Halsted. This station's impact is visible at distance bin 21 for the Northbound direction and distance bin 43 for the Southbound direction, where changes in headway patterns and propagation are evident. These specific shifts in the headway distribution are due to southbound trains switching to northbound service.

The processed data, with headway values normalized to the interval [0, 1] using min-max scaling, were organized into input-output sequence pairs. Each input sequence captures 30 minutes of historical headway data across all spatial bins (on a minute by minute basis), represented as a tensor $X \in \mathbb{R}^{L \times N_d \times N_{\text{dir}} \times 1}$, where $L$ denotes the number of historical time steps, $N_d$ the number of distance bins, and $N_{\text{dir}}$ the number of directional bins (e.g., Northbound and Southbound). A novel input, the future terminal headways, was incorporated separately, capturing the scheduled headway at the terminal station(s) over the prediction horizon. The target output, denoted by $Y \in \mathbb{R}^{F \times N_d \times N_{\text{dir}} \times 1}$, comprises the predicted headway values for all stations across the prediction horizon, where $F$ represents the number of future time steps, which is set to 15 minutes.

The proposed model was trained using the Adam optimizer with a learning rate of 0.001 and MSE as the loss function \cite{kingma2014adam}. Training spanned 100 epochs, with a batch size of 32, leveraging a large-scale dataset of historical headway data ($X \in \mathbb{R}^{L \times N_d \times N_{\text{dir}} \times 1}$) and target terminal schedules ($T \in \mathbb{R}^{F \times N_{\text{dir}} \times 1}$). A held-out validation set, comprising 20\% of the data, is used to monitor for overfitting, with training and validation loss convergence shown in Figure~\ref{fig:fig_loss}. Early stopping was employed to halt training if validation loss does not improve for 50 epochs. This training process enabled the model to capture spatiotemporal dependencies, producing reliable headway predictions ($\hat{\mathbf{Y}} \in \mathbb{R}^{F \times N_d \times N_{\text{dir}} \times 1}$) for operational use.

Key hyperparameters and their chosen values are summarized in Table \ref{tab:model_params}.

\begin{table}[H]
\centering
\caption{Model Hyperparameters and Training Configuration}
\label{tab:model_params}
\begin{tabular}{|l|l|}
\hline
\textbf{Parameter} & \textbf{Value} \\
\hline
Input Lookback Steps & 30 minutes (30 time bins) \\
prediction Horizon (Single Step) & 15 minutes (15 time bins) \\
Recursive Prediction Horizon & Up to 60 minutes \\
Number of Distance Bins & 64 \\
Number of Directions & 2 (Northbound, Southbound) \\
ConvLSTM Filters & 32 \\
ConvLSTM Kernel Size & $(3, 1)$ \\
Activation Function & ReLU \\
Optimizer & Adam \\
Learning Rate & $0.001$ \\
Loss Function & MSE \\
Batch Size & 32 \\
\hline
\end{tabular}
\end{table}

\subsection{Model Performance}

\begin{figure}[H]
    \centering
    \includegraphics[width=0.5\linewidth]{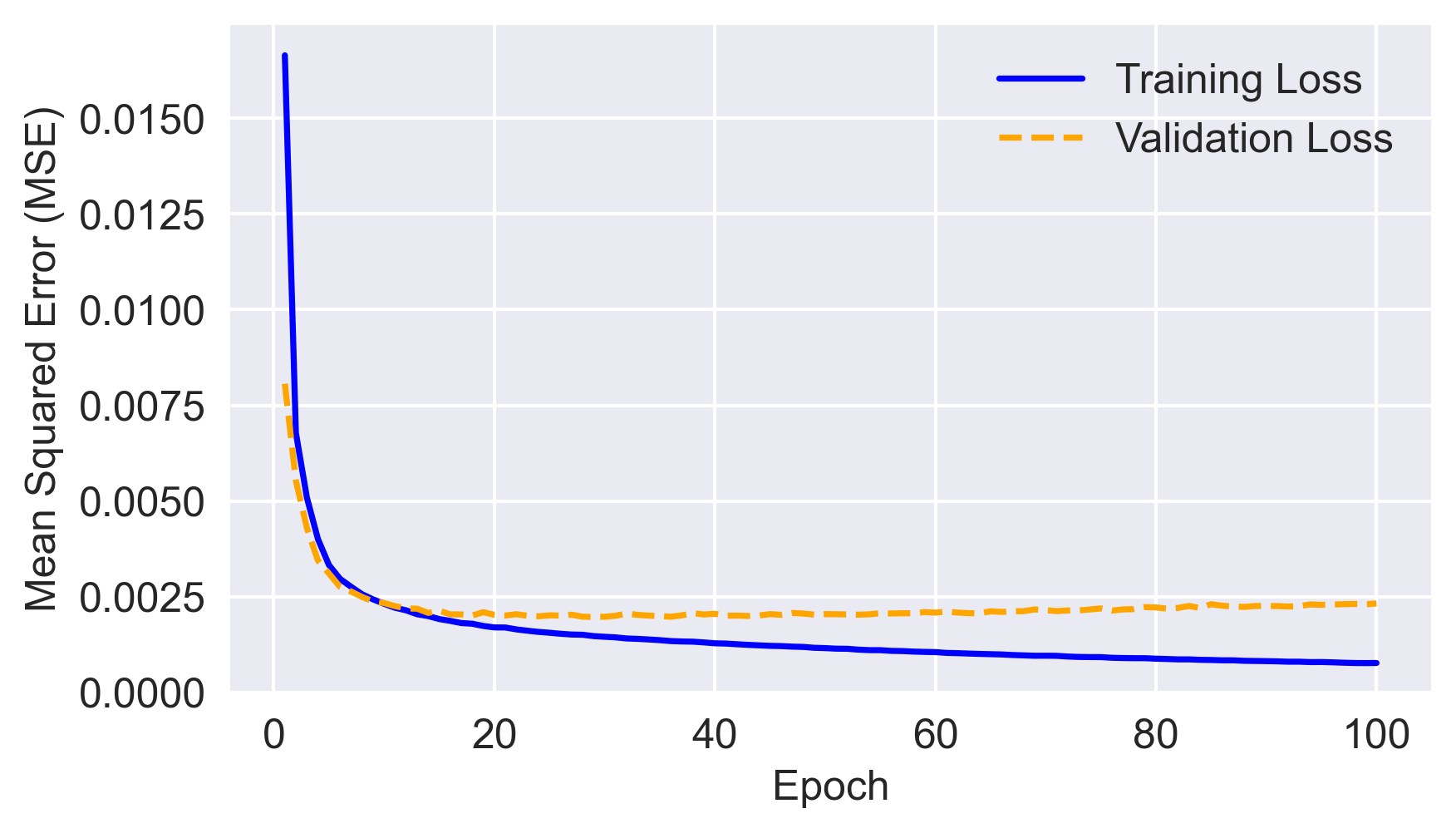}
    \caption{Training and Validation Loss over Epochs}
    \label{fig:fig_loss}
\end{figure}

The proposed model demonstrated robust performance in predicting spatiotemporal headway dynamics. The training and validation loss curves, in Figure \ref{fig:fig_loss}, show a consistent decrease and convergence, indicating effective learning and generalization. The model's ability to minimize MSE across different datasets suggests its capacity to capture the complex temporal dependencies and spatial correlations inherent in metro headway propagation.

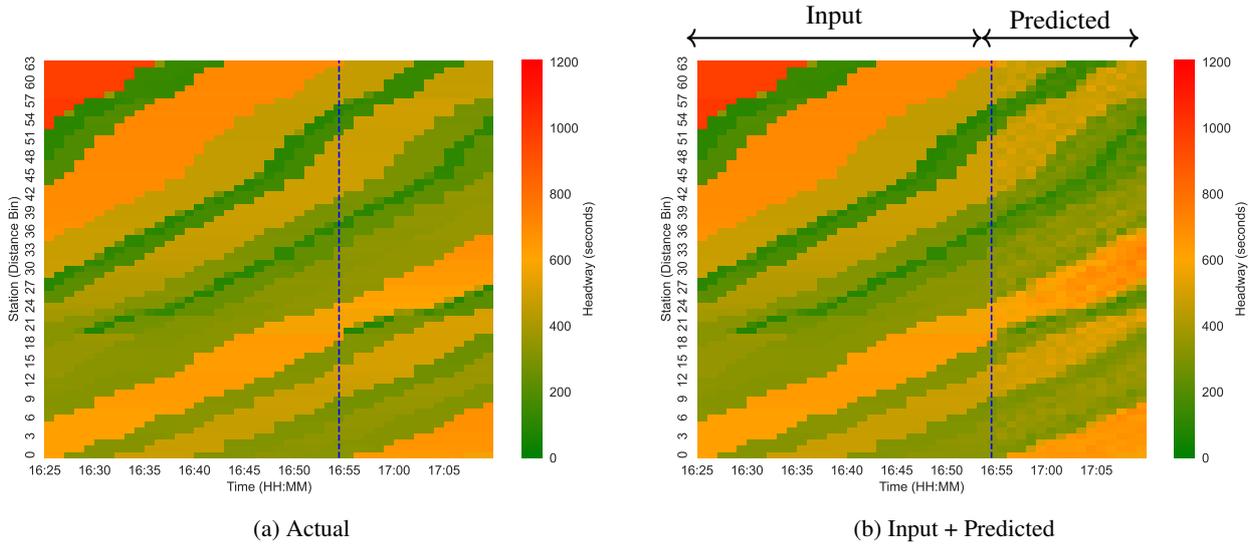
\begin{figure}[H]
\centering
\begin{subfigure}{0.48\textwidth}
    \centering
    \includegraphics[width=\textwidth]{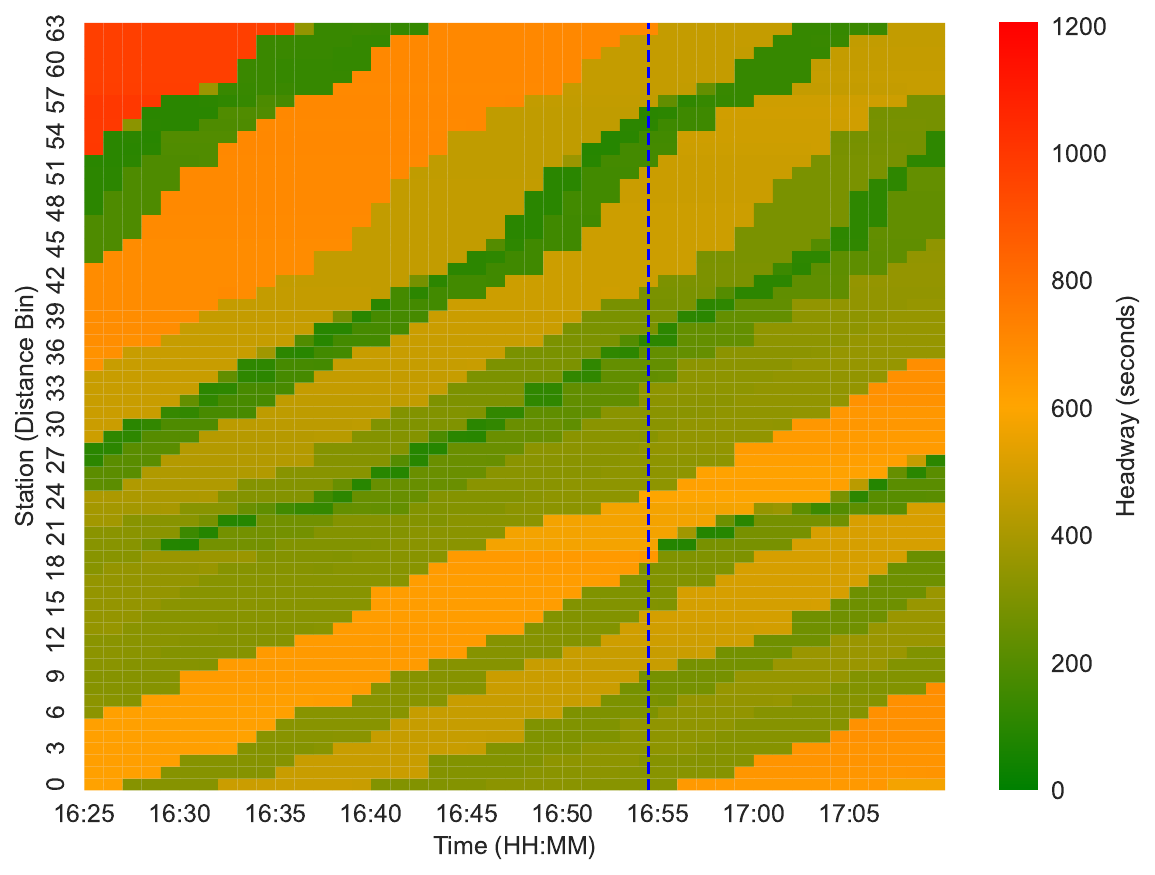}
    \caption{Actual}
    \label{fig:actual_heatmap_15min}
\end{subfigure}
\hfill
\begin{subfigure}{0.48\textwidth}
    \centering
    \includegraphics[width=\textwidth]{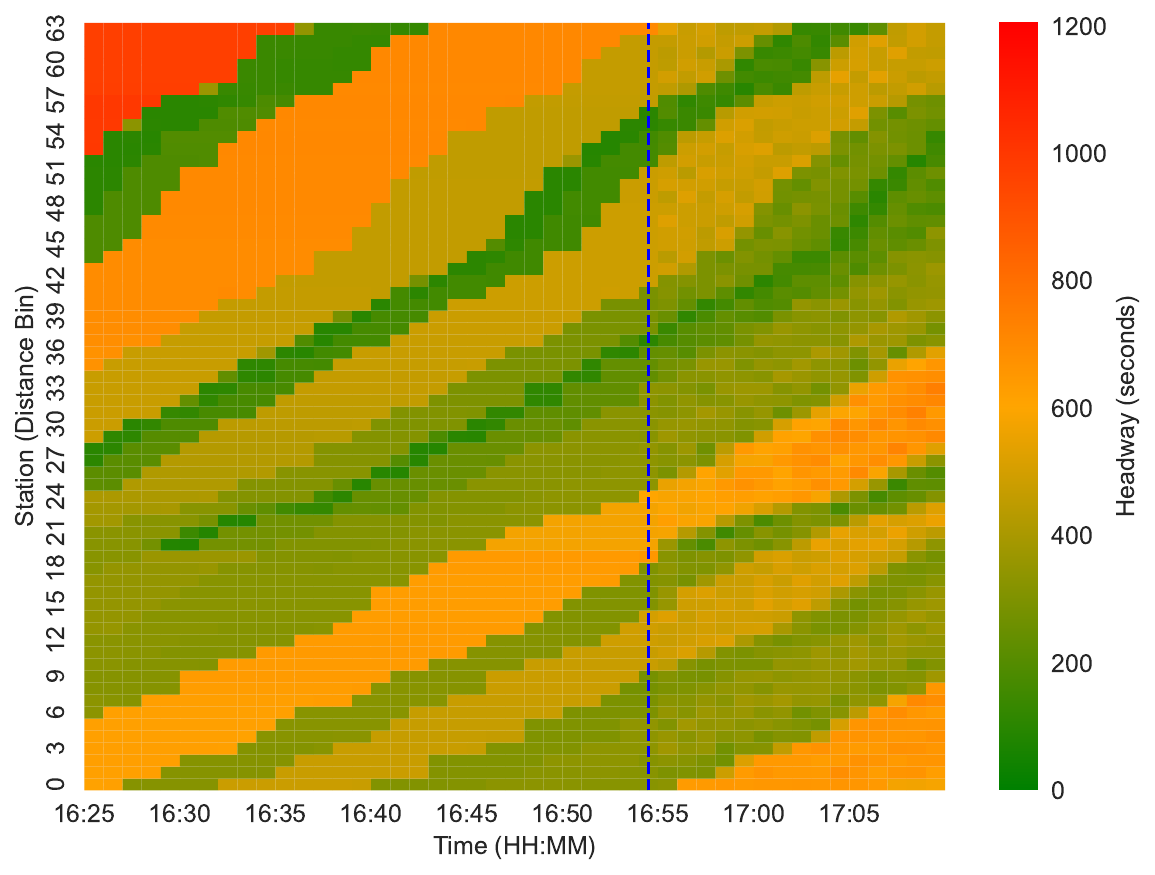}
    \caption{Input + Predicted}
    \label{fig:predicted_heatmap_15minutes}
\end{subfigure}

\begin{tikzpicture}[overlay, remember picture]
    \coordinate (center-right) at ([yshift=0.5cm]$(current page.north)-(-3.25cm,2.9cm)$);
    \draw[<->, thick] ([xshift=-0.15\textwidth]center-right) -- ([xshift=0.085\textwidth]center-right)
        node[midway, above, yshift=-0.01cm] {Input};
    \draw[<->, thick] ([xshift=0.085\textwidth]center-right) -- ([xshift=0.21\textwidth]center-right)
        node[midway, above, yshift=-0.01cm] {Predicted};
\end{tikzpicture}

\caption{Headway heatmaps for the CTA Blue Line, Northbound direction, during the afternoon peak. The plots show headways (seconds) for 30 minutes of historical data and 15 minutes of actual (a) or predicted (b) headways.}
\label{fig:prediction_heatmaps_15minutes}
\end{figure}

To evaluate the proposed model's predictive performance for the next 15-minute (minute-by-minute) prediction horizon, heatmaps were generated to compare actual and predicted headway values for the northbound direction of the Blue Line during the afternoon peak period. Figure \ref{fig:prediction_heatmaps_15minutes} illustrates the spatiotemporal distribution of headways for a replication from the dataset, combining 30 minutes of historical data with 15 minutes of actual headways. Figure (\ref{fig:actual_heatmap_15min}) shows the actual headway heatmap while Figure (\ref{fig:predicted_heatmap_15minutes}) is the predicted headway heatmap across the entire line. The vertical blue line demarcates the transition from historical to future periods. The headway patterns at distance bin 21, corresponding to the UIC-Halsted station, where southbound trains short-turn to northbound, reveal complex operational dynamics, as evidenced by pronounced headway variations. These variations highlight the model’s ability to capture complex spatiotemporal dependencies, particularly at key operational points. The close alignment between the actual and predicted heatmaps in the short-term prediction horizon underscores the model’s effectiveness in leveraging historical and terminal headway data for accurate predictions, as further supported by the RMSE values reported in Table \ref{tab:rmse_r2_predictions}.

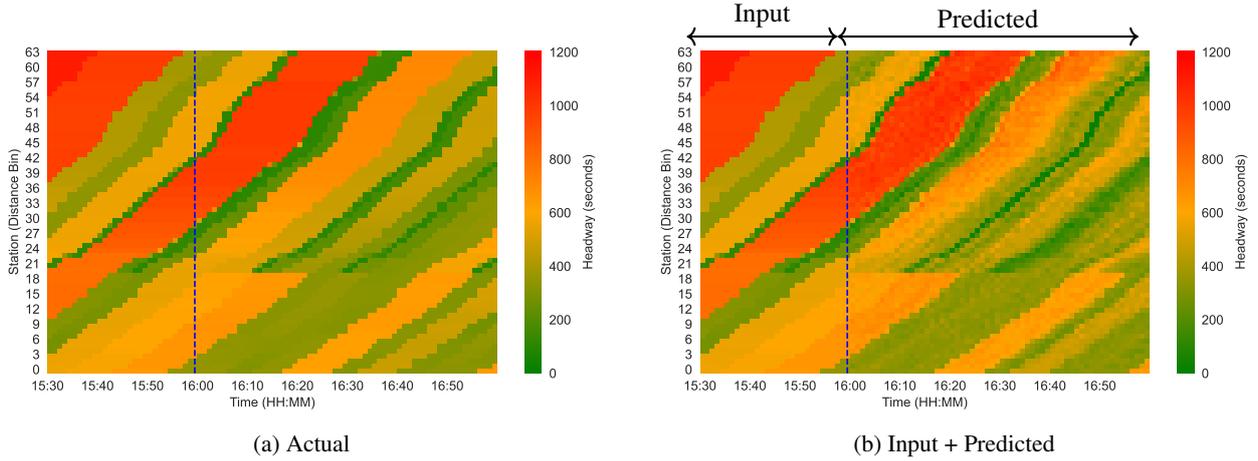
\begin{figure}[H]
\centering
\begin{subfigure}{0.48\textwidth}
    \centering
    \includegraphics[width=\textwidth]{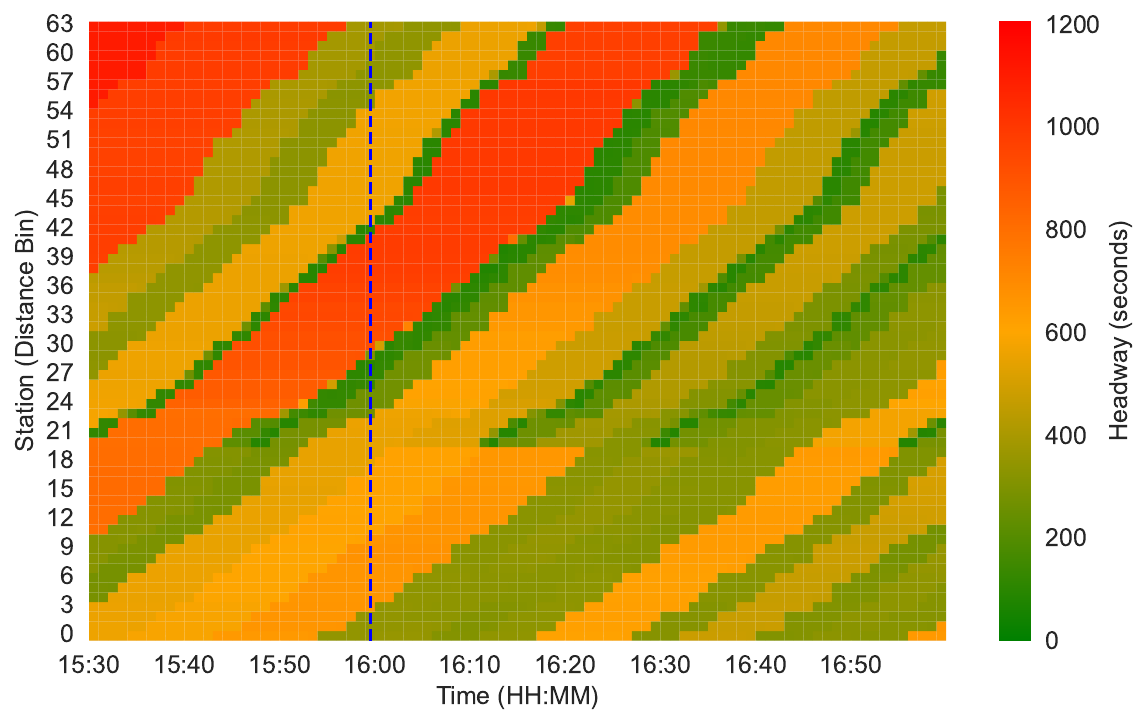}
    \caption{Actual}
    \label{fig:actual_heatmap_60min}
\end{subfigure}
\hfill
\begin{subfigure}{0.48\textwidth}
    \centering
    \includegraphics[width=\textwidth]{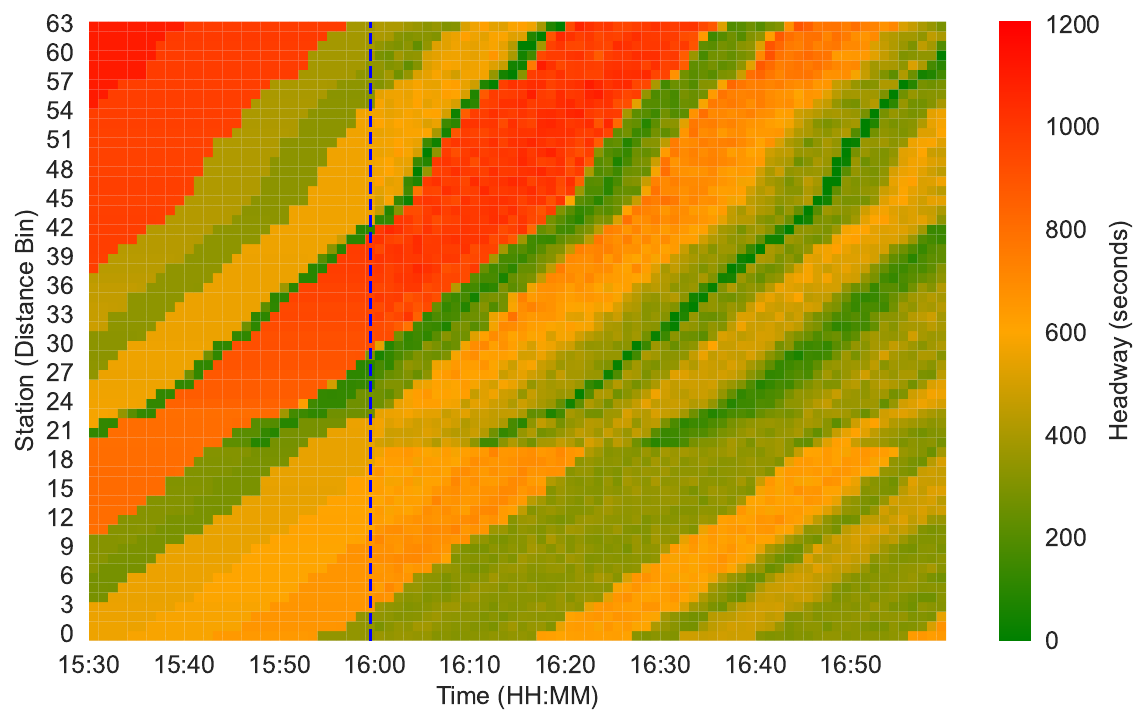}
    \caption{Input + Predicted}
    \label{fig:predicted_heatmap_60min}
\end{subfigure}

\begin{tikzpicture}[overlay, remember picture]
    \coordinate (center-right) at ([yshift=0.5cm]$(current page.north)-(-3.25cm,3.0cm)$);
    \draw[<->, thick] ([xshift=-0.15\textwidth]center-right) -- ([xshift=-0.03\textwidth]center-right)
        node[midway, above, yshift=-0.01cm] {Input};
    \draw[<->, thick] ([xshift=-0.03\textwidth]center-right) -- ([xshift=0.21\textwidth]center-right)
        node[midway, above, yshift=-0.01cm] {Predicted};
\end{tikzpicture}

\caption{Headway heatmaps for the CTA Blue Line, Northbound direction, during the afternoon peak, showing headways (seconds) across entire line for 30 minutes of historical data and a 60-minute prediction horizon for actual (a) and predicted (b) headways.}
\label{fig:prediction_heatmaps_60min}
\end{figure}

To evaluate the model’s performance across extended prediction horizons, sliding window predictions were generated recursively during the afternoon peak period, using the entire dataset. Figure \ref{fig:prediction_heatmaps_60min} illustrates the spatiotemporal distribution of headways for a replication, combining 30 minutes of historical data with 60 minutes of actual (\ref{fig:actual_heatmap_60min}) and predicted (\ref{fig:predicted_heatmap_60min}) headways across the entire line.

\newpage

\begin{figure}[H]
\centering
\begin{subfigure}{0.48\textwidth}
    \centering
    \includegraphics[width=\textwidth]{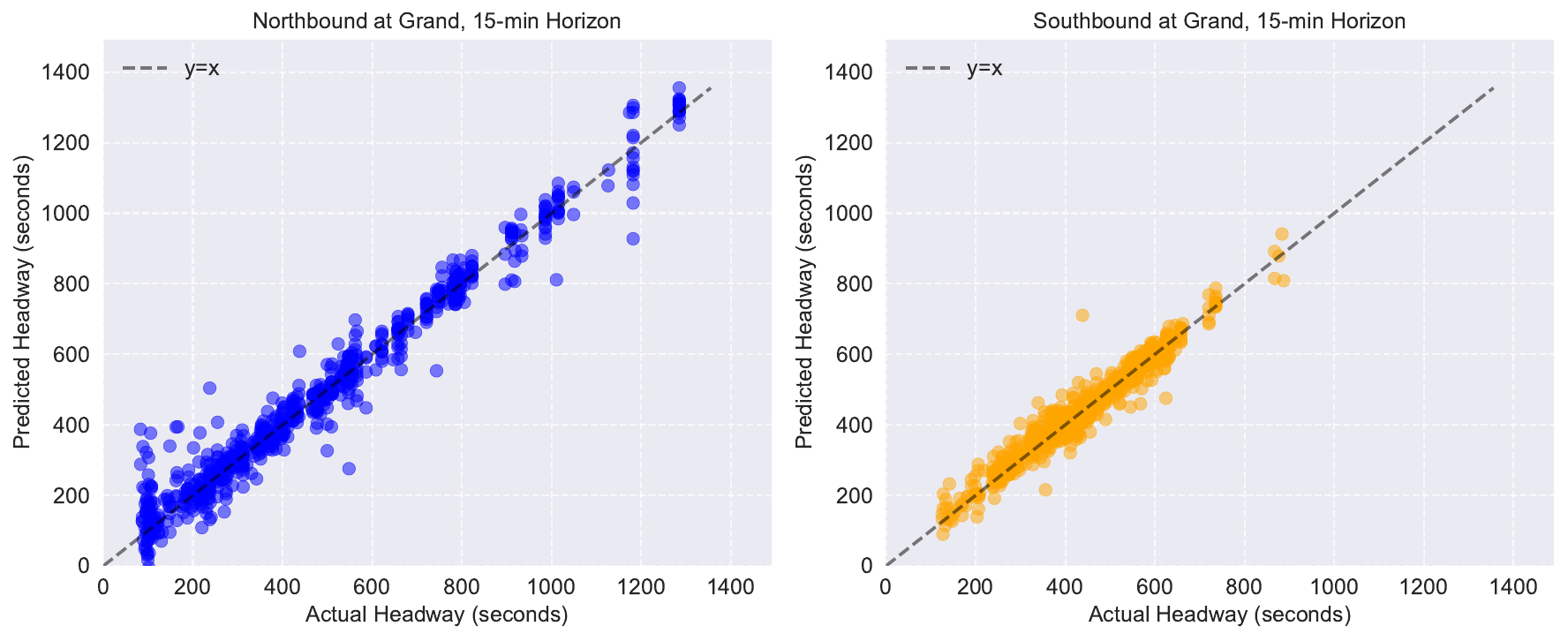}
    \caption{15-minute horizon}
    \label{fig:scatter_horizon_15}
\end{subfigure}
\hfill
\begin{subfigure}{0.48\textwidth}
    \centering
    \includegraphics[width=\textwidth]{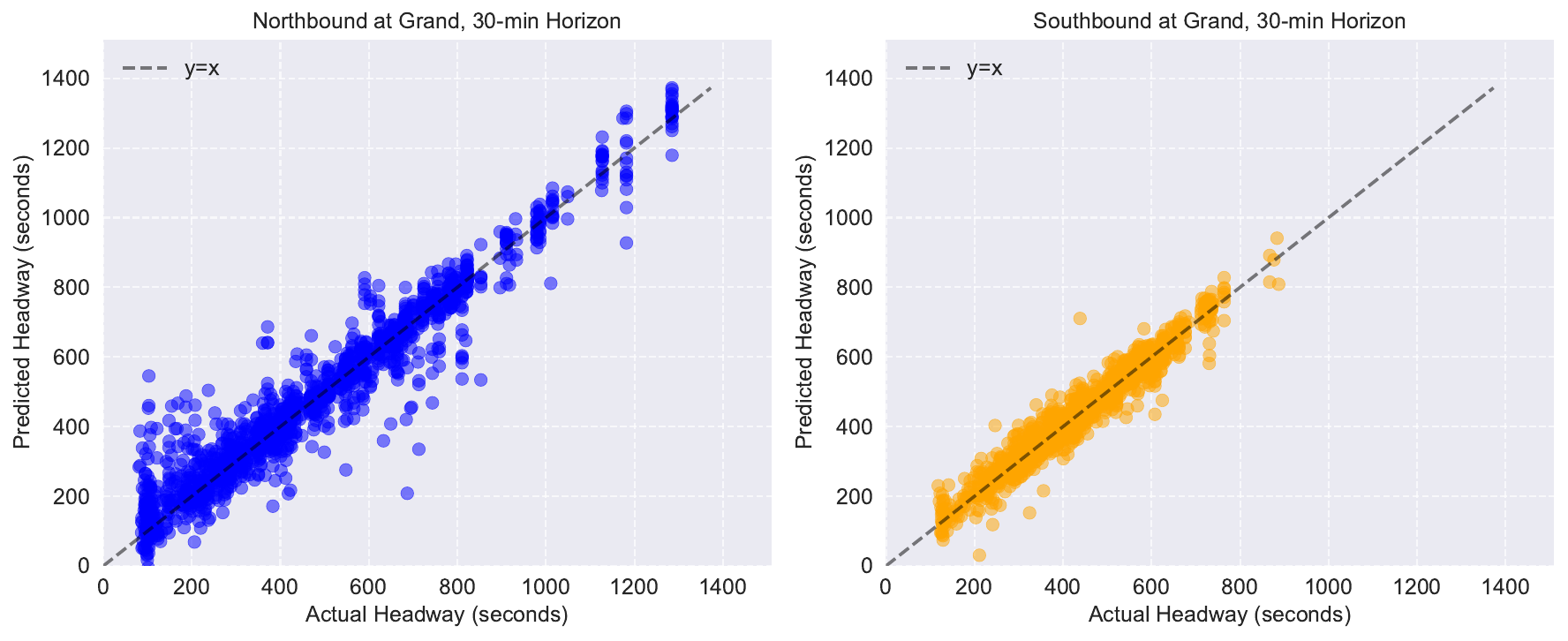}
    \caption{30-minute horizon}
    \label{fig:scatter_horizon_30}
\end{subfigure}

\vspace{0.2cm}

\begin{subfigure}{0.48\textwidth}
    \centering
    \includegraphics[width=\textwidth]{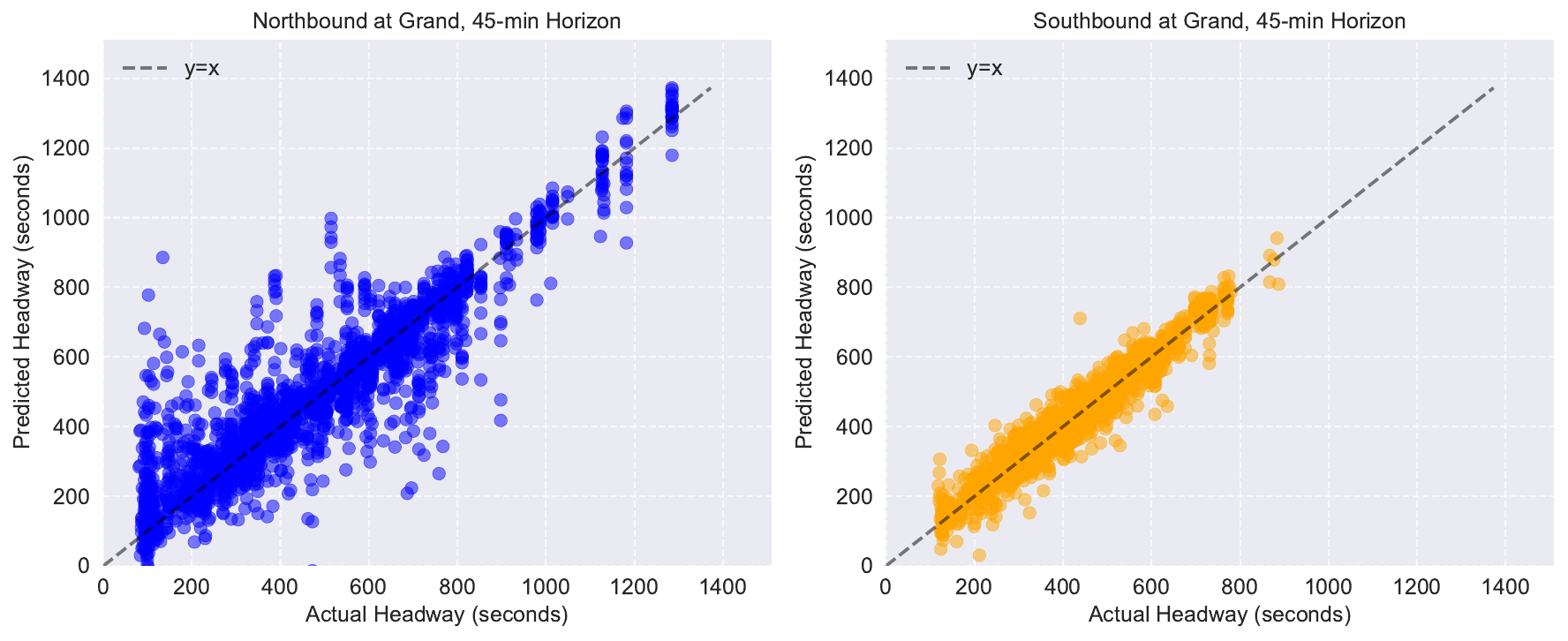}
    \caption{45-minute horizon}
    \label{fig:scatter_horizon_45}
\end{subfigure}
\hfill
\begin{subfigure}{0.48\textwidth}
    \centering
    \includegraphics[width=\textwidth]{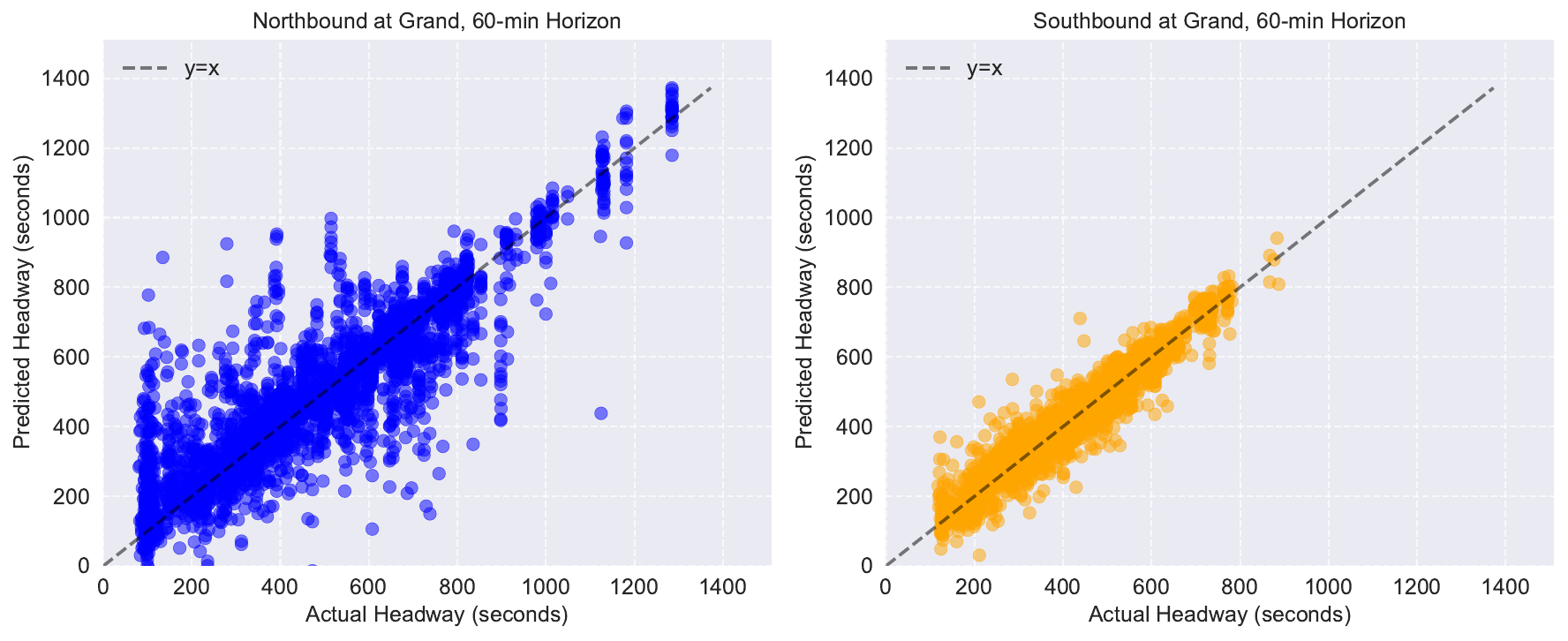}
    \caption{60-minute horizon}
    \label{fig:scatter_horizon_60}
\end{subfigure}

\caption{Scatter plots of actual vs. predicted headways, with Northbound and Southbound directions for 15, 30, 45, and 60-minute prediction horizons at Grand Station.}
\label{fig:scatter_headways}
\end{figure}

Figure~\ref{fig:scatter_headways} presents scatter plots comparing actual and predicted headways at Grand station for the Northbound and Southbound  directions across four prediction horizons: 15, 30, 45, and 60 minutes. The plots aggregate headway predictions across all replications, providing a comprehensive view of model performance over varying time horizons. For the 15-minute horizon, the average RMSE is 48 and 28 seconds for the Northbound and Southbound directions, respectively. For the 30-minute horizon, it increases to 68 seconds for Northbound) and 32 seconds for Southbound. The 45-minute horizon yields an RMSE of 100 seconds for Northbound and 35 seconds for Southbound, while the 60-minute horizon shows 116 seconds for Northbound and 40 seconds for Southbound directions. These metrics indicate better predictive accuracy for Southbound headways across all horizons, with errors increasing as the prediction horizon extends.

Table \ref{tab:rmse_r2_predictions} reports the average RMSE and $R^2$ for headway predictions across all replications for both northbound and southbound directions of the line, covering 15, 30, 45, and 60-minute prediction horizons, with RMSE values ranging from 55 seconds to 110 seconds and $R^2$ values from 0.98 to 0.79. The increasing RMSE and decreasing $R^2$ with longer horizons are expected, as predictions for 45 and 60-minute horizons rely on a sliding window approach, where predicted headways are recursively fed back into the model as inputs, introducing cumulative errors. Notably, southbound predictions consistently exhibit slightly lower RMSE and higher $R^2$ compared to northbound, potentially due to operational dynamics at UIC-Halsted, where short-turning trains from southbound to northbound are moved to the middle track, which potentially creates conflicts with full-service trains from Forest Park. Despite higher errors at the 60-minute horizon, the RMSE remains below 2 minutes for both directions, which is promising for maintaining service reliability through control strategies at terminal stations.

\begin{table}[H]
	\caption{Performance metrics for headway predictions across prediction horizons and directions}\label{tab:rmse_r2_predictions}
	\begin{center}
		\begin{tabular}{l c c c}
			Direction & Horizon (minutes) & RMSE (seconds) & $R^2$ \\\hline
			Northbound & 15 & 55 & 0.95 \\
			Northbound & 30 & 72 & 0.91 \\
			Northbound & 45 & 94 & 0.86 \\
			Northbound & 60 & 110 & 0.79 \\
			Southbound & 15 & 47 & 0.98 \\
			Southbound & 30 & 70 & 0.93 \\
			Southbound & 45 & 86 & 0.87 \\
			Southbound & 60 & 100 & 0.82 \\\hline
		\end{tabular}
	\end{center}
\end{table}

To highlight the practical utility of the proposed spatiotemporal headway prediction framework, particularly its capacity to assist dispatchers in real-time operational decision-making, we conducted an experiment by making changes to the terminal headways. This capability is important for managing the complex dynamics of urban rail lines, where factors like dynamic passenger loads and complex short-turning operations introduce significant headway variability. It shows the model's ability to act as a "what-if" tool, allowing dispatchers to quickly evaluate the future impact of their immediate dispatching choices.

\begin{figure}[htbp]
  \centering
  \begin{subfigure}[t]{0.32\textwidth}
    \centering
    \includegraphics[width=\textwidth]{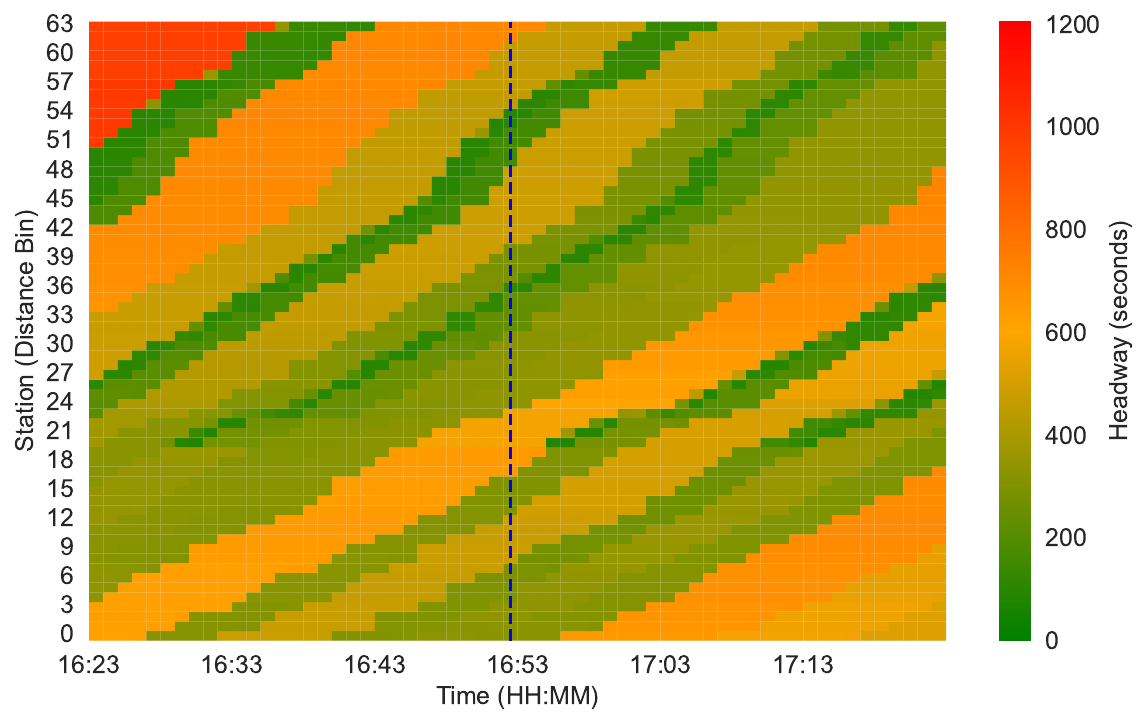}
    \caption{Actual}
    \label{fig:actual_headways}
  \end{subfigure}
  \hfill
  \begin{subfigure}[t]{0.32\textwidth}
    \centering
    \includegraphics[width=\textwidth]{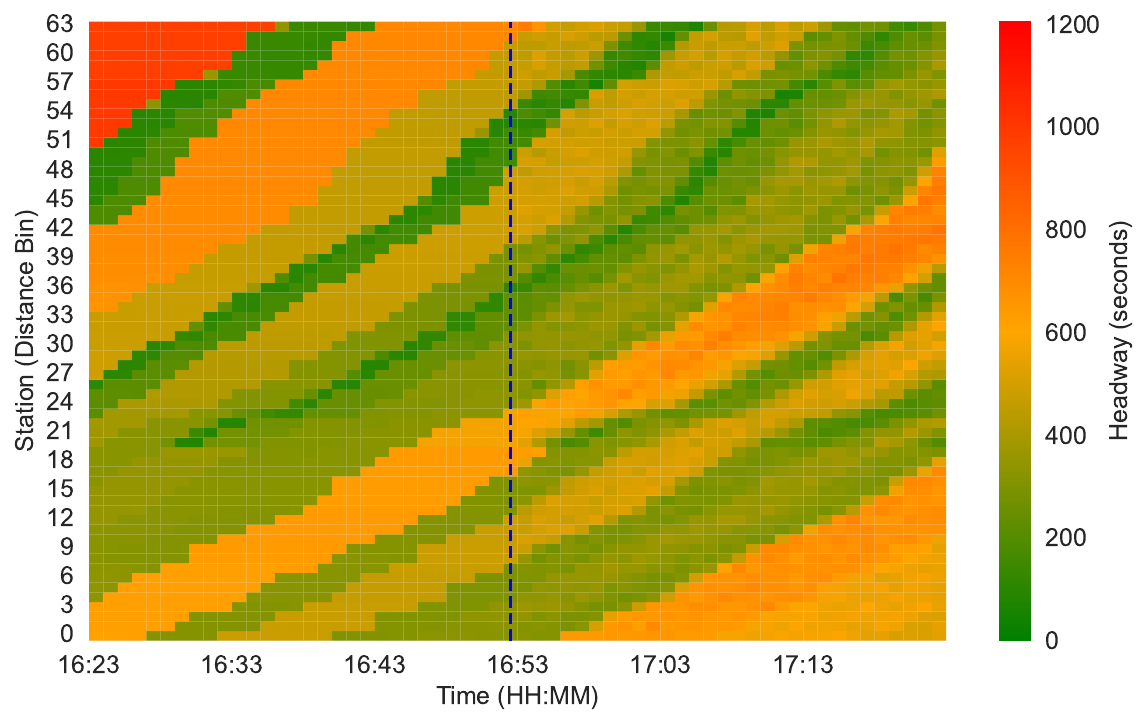}
    \caption{Input + Predicted without Terminal Headway Adjustment}
    \label{fig:predicted_actual_headways}
  \end{subfigure}
  \hfill
  \begin{subfigure}[t]{0.32\textwidth}
    \centering
    \includegraphics[width=\textwidth]{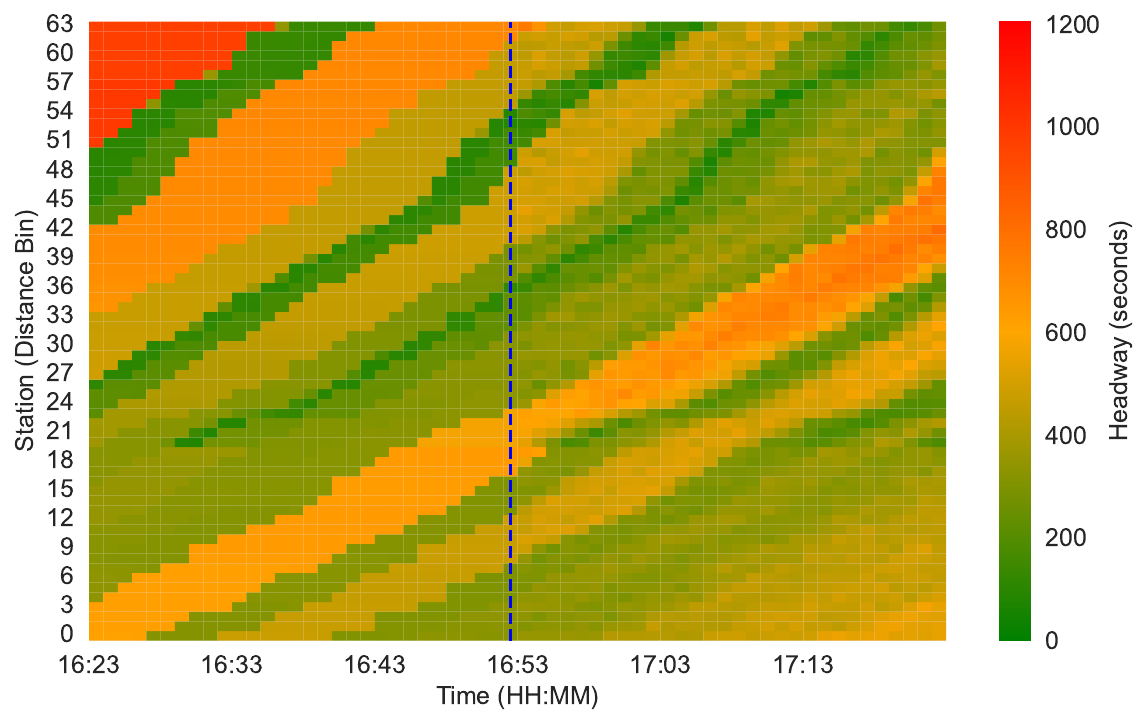}
    \caption{Input + Predicted with Terminal Headway Adjustment}
    \label{fig:predicted_adjusted_headways}
  \end{subfigure}
  \caption{Headway propagation for Northbound  showing (a) actual headways, (b) predicted headways using actual terminal headways, and (c) predicted headways using adjusted Northbound terminal headways. Headway values are in seconds, with color intensity indicating magnitude.}
  \label{fig:headway_comparison}
\end{figure}

Figure~\ref{fig:headway_comparison} illustrates the propagation of headways for the Northbound direction from terminal to the downstream stations over a 30-minute period, enabling rapid visualization of the impact of dispatching decisions on service reliability. Figure (\ref{fig:actual_headways}) shows 30 minutes of historical headway data followed by 30 minutes of actual headways, serving as the baseline for service performance. Figure (\ref{fig:predicted_actual_headways}) combines the same historical data with 30 minutes of predicted headways using actual terminal headways, reflecting model predictions without adjustments. Figure (\ref{fig:predicted_adjusted_headways}) presents predicted headways using adjusted terminal headways, incorporating the holding back strategy, where dispatchers hold trains to achieve more uniform headways. The comparison highlights how adjusted terminal headways reduce headway variability downstream, enhancing service reliability.

\section{Conclusions}

This study developed and evaluated a ConvLSTM-based framework for spatiotemporal headway prediction on the CTA Blue Line. The results, using simulated data, show robust performance across 15 to 60-minute prediction horizons. The proposed model maintains high accuracy for short-term predictions and practical accuracy at longer horizons, despite short-turning operations at UIC-Halsted. Results highlight the model's ability to capture headway dynamics at stations like Cicero and Grand, with Southbound predictions outperforming Northbound due to reduced operational conflicts. The incorporation of planned terminal headways enhances the model's utility for proactive dispatching, enabling transit agencies to optimize service reliability during peak periods.

Future research should focus on several directions to enhance the framework's applicability of the approach. First, integrating real-time AVL data and passenger volume inputs could improve prediction accuracy, particularly at high-traffic stations like Grand, where operational variability is higher. Second, exploring more advanced NN architectures, such as attention-based models, may better capture complex spatiotemporal dependencies and mitigate errors in recursive predictions. Finally, validating the model across other metro systems with diverse operational characteristics would strengthen its generalizability.




\section{Conflict of Interest Statement}
The authors declared no potential conflicts of interest with respect to the research, authorship, and/or publication of this article.

\newpage

\bibliographystyle{unsrt}  
\bibliography{references}  



\end{document}